\documentclass[runningheads]{llncs}
\usepackage{times}
\usepackage{latexsym}
\usepackage{amsmath,graphicx,hyperref}
\usepackage{multirow} 
\usepackage{booktabs}
\usepackage{algorithmic}
\usepackage{textcomp}
\usepackage{xcolor}
\usepackage{multirow} 
\usepackage{amsfonts}
\usepackage{makecell} 
\usepackage[T1]{fontenc}
\usepackage{cite}
\usepackage{graphicx}
\begin{document}
\title{Evaluating Multimodal Large Language Models on Spoken Sarcasm Detection} 
%
%
\author{Zhu Li\orcidID{0000-0002-1409-2482} \and
Xiyuan Gao\orcidID{0000-0003-0870-6721} \and 
Yuqing Zhang\orcidID{0000-0002-4362-103X} \and
Shekhar Nayak\orcidID{0000-0002-4277-4851} \and \\
Matt Coler\orcidID{0000-0002-7631-5063}}
\authorrunning{Z. Li et al.}
%
\institute{University of Groningen, Broerstraat 5, 9712 CP, The Netherlands \\
\email{\{zhu.li, xiyuan.gao, yuqing.zhang, s.nayak, m.coler\}@rug.nl}}
\maketitle              
\begin{abstract}
Sarcasm detection remains challenging because sarcastic intent often depends on information across text, speech, and vision. We systematically evaluate large language models (LLMs) and multimodal LLMs (MLLMs) for sarcasm detection on English (MUStARD++) and Chinese (MCSD), considering both feature-extraction settings and direct classification. In the feature-extraction settings, audio provides the strongest unimodal performance on both datasets, while multimodal fusion shows dataset-dependent benefits. In direct classification, Qwen2.5-Omni achieves the strongest zero-shot performance among the evaluated models, with further substantial gains after fine-tuning. Error analysis further shows that additional modalities can both correct and introduce errors, highlighting the need for selective multimodal integration.

\keywords{Sarcasm detection \and multimodal understanding \and large language models \and zero-shot learning \and multimodal fusion}
\end{abstract}
\section{Introduction}
\label{sec:intro}

Sarcasm is a complex and pervasive aspect of human communication, where the intended meaning diverges from the literal expression.  
While sarcasm can be conveyed through linguistic cues, explicit markers are often absent. Detecting sarcastic intent often requires additional information, such as prosodic, facial, and gestural cues (e.g., overemphasis on a word or an exaggerated facial expression) \cite{li2024functional}. Moreover, recognizing sarcasm frequently depends on detecting contextual incongruity across modalities. 
A speaker who says ``Oh, that's just great'' in a flat voice while rolling her eyes uses positive words, but her prosody and gesture signal sarcasm \cite{castro-etal-2019-towards}. Detecting the mismatch requires more than the transcript.
Beyond inter-modal interaction, sarcasm is also deeply shaped by cultural context \cite{joshi2016cultural}. 
The production and perception of sarcasm vary across languages, yet most multimodal sarcasm research uses English data. Systems built on it may not generalize to other languages and cultures.
These observations highlight the importance of developing multimodal sarcasm detection systems capable of capturing cross-modal incongruities while adapting to linguistic and cultural diversity.

We evaluate unimodal and multimodal models, alone and in combination, on two sarcasm datasets: MUStARD++ \cite{ray-etal-2022-multimodal}, drawn from English TV sitcoms, and MCSD \cite{gao2025multimodal}, drawn from Chinese stand-up comedy.
We test the zero-shot, few-shot, and LoRA fine-tuned performance of LLMs and MLLMs, and investigate their role as feature extractors within a collaborative gating-based fusion framework \cite{liu2019use}. 
We extend the evaluation from English to Chinese, comparing the contributions of text, audio, and visual inputs on both datasets\footnote{Data and code are available at \url{https://github.com/Abel1802/NCMMSC26}}. Because the two corpora differ in genre as well as language, we treat the comparison as bilingual as opposed to cross-lingual.

\section{Related Work}
\label{sec:related}
Early research has primarily focused on \textit{visual-textual sarcasm detection}, where images and captions jointly convey sarcastic meaning \cite{schifanella2016detecting}. Such initial work often used separate encoders 
for each modality and explored increasingly sophisticated multimodal fusion techniques, 
ranging from simple concatenation \cite{schifanella2016detecting} to attention-based modeling of inter- and intra-modal incongruities \cite{pan2020modeling}. More recent approaches employ multimodal encoders 
such as VisualBERT 
and CLIP \cite{radford2021learning}, or integrate large language models (LLMs) via prompt engineering for sarcasm detection \cite{ding2022multi}.
Despite these advances, research has largely focused on text-only or visual-textual scenarios, leaving sarcasm in natural spoken interactions underexplored. 

In contrast, sarcasm in videos (e.g., sitcoms and stand-up comedy) requires reasoning across speech, facial expressions, gestures, and textual transcripts. Prior work on datasets such as MUStARD \cite{castro-etal-2019-towards} and MUStARD++ \cite{ray-etal-2022-multimodal} demonstrated that audio and video cues significantly enhance sarcasm detection. Recent speech-focused studies have further investigated how semantic and prosodic cues contribute to perceived sarcasm, including controlled speech synthesis and perception-based analyses \cite{li2026modeling,li2026zerotts}.
Subsequent methods leveraged multimodal attention \cite{wu2021modeling, aggarwal2023multimodal}, optimal-transport alignment \cite{pramanick2022multimodal}, 
or collaborative gating fusion \cite{ray-etal-2022-multimodal, raghuvanshi2025intra} to improve the overall performance. 
However, these approaches primarily rely on task-specific architectures for feature fusion, without systematically assessing the potential of recent multimodal LLMs (MLLMs). 

LLMs have also recently been explored for supporting sarcastic speech annotation, showing their potential for constructing and analyzing spoken sarcasm datasets \cite{li2025leveraging}. Meanwhile, general-purpose MLLMs such as Qwen2.5-Omni have achieved impressive reasoning capabilities across text, audio, and visual modalities \cite{Qwen2.5-Omni}. Their multimodal understanding capabilities open new opportunities for tasks involving complex cross-modal incongruities. However, the role of MLLMs in multimodal sarcasm detection for spoken interactions remains largely unexplored. It remains unclear whether the emergent multimodal reasoning abilities of MLLMs extend to such fine-grained pragmatic phenomena.
Existing LLM benchmarks either restrict multimodal sarcasm to image-text memes, as GOAT does \cite{lin2024goat}, or treat it as one task among many, as MM-BigBench does \cite{yang2023mm}. Recent studies have begun to test conversational sarcasm across text, audio, and video. 
Complementary to Chen et al. \cite{chen2026hear}, who diagnose which acoustic properties drive sarcasm-related errors, we focus on modality-conditioned MLLM performance through feature-fusion comparisons and modality-level error-transition analysis. SarcasmMiner \cite{li2026sarcasmminer} instead studies post-training for multimodal sarcasm reasoning, whereas our work is evaluation-focused.

\begin{figure*}[t]
    \centering
    \includegraphics[width=\linewidth]{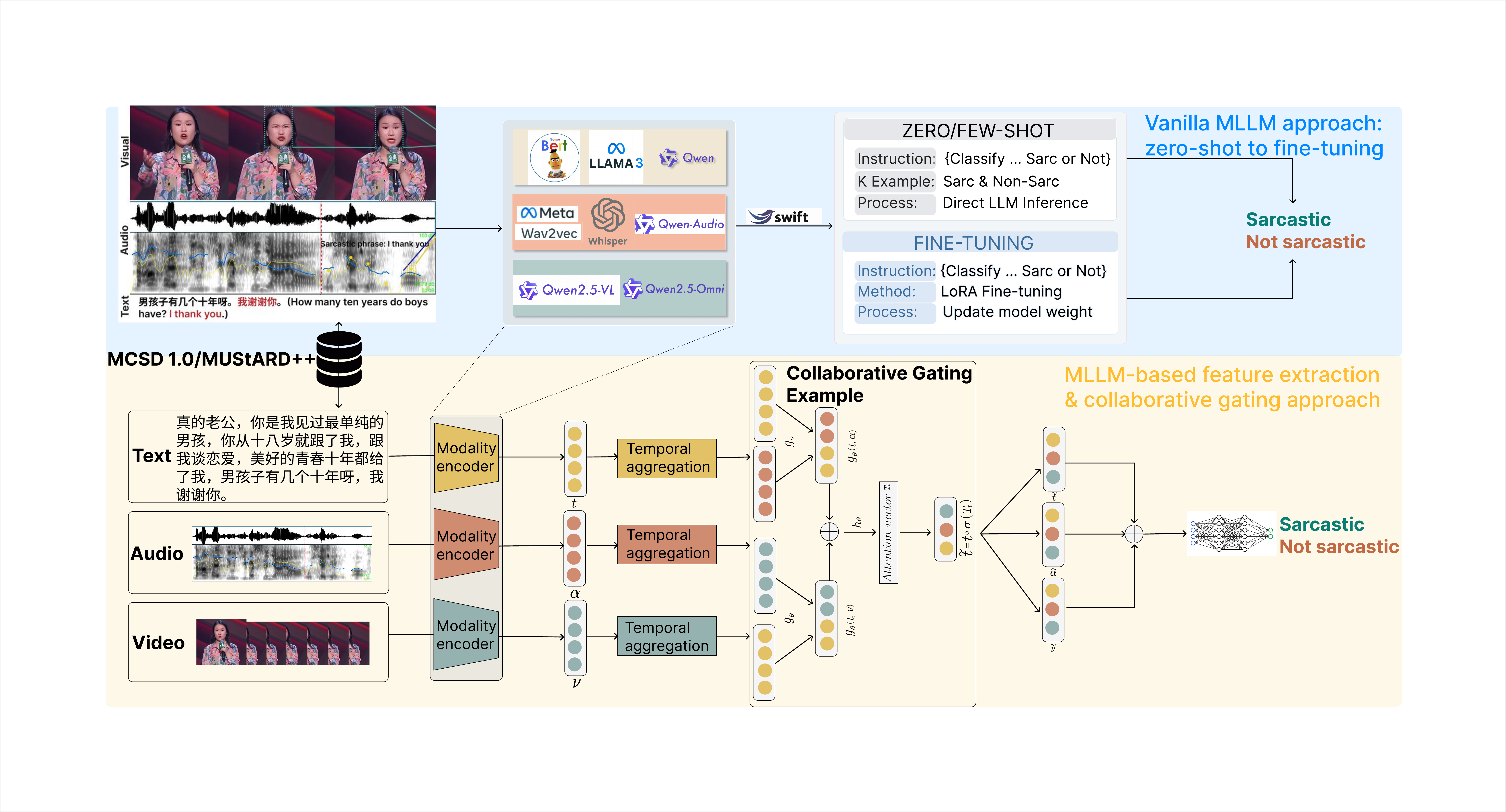}
    \caption{Overview of our framework.}
    \label{fig:arch}
\end{figure*}

\section{Method}

Figure~\ref{fig:arch} illustrates the overall framework. We begin by evaluating recent LLMs and MLLMs on sarcasm detection under zero-shot, few-shot, and LoRA fine-tuning conditions. While they exhibit promising capabilities especially after fine-tuning, their performance still leaves room for improvement. 
Motivated by this, we explore a second approach that treats LLMs and MLLMs as feature extractors, fusing their output features via a collaborative gating module.

\subsection{MLLM-based Sarcasm Detection: Prompting and Fine-Tuning}
We evaluate a mix of unimodal and multimodal large models, comparing their performance in zero-shot, few-shot, and LoRA fine-tuning sarcasm detection settings. Our evaluation covers LLaMA 3 \cite{dubey2024llama3} (text-only), Qwen2-Audio \cite{chu2024qwen2} (audio-language), Qwen2.5-VL \cite{bai2025qwen25vl} (vision–language), and Qwen2.5-Omni \cite{Qwen2.5-Omni} (audio, vision, language). 

We evaluate three setups. In \textbf{zero-shot prompting}, each model receives task instructions and every modality it accepts, but no labeled examples. In \textbf{few-shot prompting}, the prompt also includes $k \in {2,4,6}$ labeled demonstrations, sampled at random from the training split with equal numbers of sarcastic and non-sarcastic items; all models receive the same demonstrations. In \textbf{fine-tuning}, each model is adapted to each dataset with LoRA \cite{hu2022lora}, providing a supervised reference point for the prompting results.


\subsection{Models as Feature Extractors for Sarcasm Detection}
\label{sec:extractor}
We also compare the effectiveness of small-scale pretrained language models versus large-scale LLMs and MLLMs as feature extractors.
For each modality, we select a conventional backbone \textbf{(Base)} and a larger-scale foundation model \textbf{(Large)}: BERT vs. LLaMA 3 for text (T), Wav2Vec 2.0 vs. Qwen2-Audio for audio (A), and ResNet50 vs. Qwen2.5-VL for video (V) \footnote{Although named `Qwen2-Audio' and `Qwen2.5-VL', both models are trained with large-scale text corpora in addition to audio/video, and thus leverage cross-modal semantic knowledge \cite{chu2024qwen2, bai2025qwen25vl}.}. 

\textbf{Text Modality:} 
The Base text encoder is BERT \cite{devlin2019bert}. We average its 768-dimensional token states to obtain an utterance embedding $t \in \mathbb{R}^{768}$.
The Large text encoder is LLaMA 3-8B \cite{dubey2024llama3}. We average its final-layer token embeddings (4096-dim) across the sequence to obtain an utterance representation.

\textbf{Audio Modality:}
For acoustic modeling, we use the pretrained Wav2Vec 2.0 \emph{Base} encoder \cite{baevski2020wav2vec2}. It generates contextualized embeddings of dimension 768, which are averaged across time, resulting in audio features $f_a \in \mathbb{R}^{768}$.  
Qwen2-Audio (7B) is adopted as a large multimodal audio-language model \cite{chu2024qwen2}. It outputs 4096-d contextual embeddings, offering a more expressive representation of acoustic content compared to wav2vec 2.0.

\textbf{Video Modality:}
For visual feature extraction, we sample $N_v=8$ keyframes from each video and process them with pretrained ResNet50 \cite{he2016resnet}. The 2048-d pooler outputs from each frame are stacked to form the visual representation matrix.
As a stronger vision-language model, Qwen2.5-VL (7B) encodes visual content jointly with textual prompts \cite{bai2025qwen25vl}. We extract 3584-dim embeddings from its vision encoder, which provide semantically enriched visual features that complement the ResNet baseline.









\textbf{Collaborative Gating Fusion.}
To integrate multiple modalities, we follow the collaborative gating framework \cite{ray-etal-2022-multimodal, liu2019use}. The modality-specific representations are first projected into a common embedding space and temporally aggregated, yielding text, audio, and visual representations $t$, $a$, and $v$, respectively.

Taking the text representation as an example, the model computes its pairwise relations with the audio and visual representations:
\[
g_{\theta}(t,a), \qquad g_{\theta}(t,v).
\]
These cross-modal relations are aggregated and transformed by $h_{\theta}$ to produce an attention vector $T_t$. The text representation is then modulated through element-wise gating:
\[
\tilde{t}=t\circ\sigma(T_t),
\]
where $\circ$ denotes element-wise multiplication and $\sigma$ is the sigmoid function.
This design allows the fusion module to adaptively modulate each modality using information from the other modalities. As illustrated in Figure~\ref{fig:arch}, taking text as an example, its representation is modulated by cross-modal relations with the audio and visual representations. The same procedure is applied to the audio and visual modalities, yielding $\tilde{a}$ and $\tilde{v}$. The resulting gated representations are then fused as
\[
h_{\mathrm{fusion}}
=
\tilde{t}\oplus\tilde{a}\oplus\tilde{v},
\]
and passed to the classifier for sarcasm detection.


\section{Experiments and Results}
\subsection{Dataset}
We evaluate our models on two datasets: 1) \textbf{MCSD} \cite{gao2025multimodal}: a recently released Multimodal Chinese Sarcasm Dataset collected from Chinese stand-up comedy. It contains 2,705 items (1,350 sarcastic and 1,355 non-sarcastic), each with aligned video, audio, and manually transcribed utterances annotated for sarcasm. We adopt a fixed 70:15:15 split, with 1,893 training, 406 validation, and 406 test samples. 
2) \textbf{MUStARD++} \cite{ray-etal-2022-multimodal}: an English multimodal sarcasm detection dataset consisting of text, speech, and video clips from TV dialogues. It contains 1,202 labeled utterances, split into 841 training, 180 validation, and 181 test examples. 

\subsection{Experimental Setup}
We follow the experimental setup of MUStARD++ \cite{ray-etal-2022-multimodal} for sarcasm detection\footnote{\url{https://github.com/cfiltnlp/MUStARD_Plus_Plus}}. Hyperparameters are tuned via grid search, with dropout rates selected from $\{0.2, 0.3, 0.4\}$, learning rates from $\{0.001,0.0001\}$, and batch sizes from $\{32, 64, 128\}$. We experiment with shared embedding sizes of $\{1024, 2048, 4096\}$ and projection embedding sizes of $\{256, 1024\}$. During fine-tuning, we set the expansion factor for the LoRA parameters to 8, and the learning rate to 1e-4 \footnote{\url{https://github.com/modelscope/ms-swift}}. 

\subsection{Prompting and Fine-tuning Evaluation}
Table~\ref{tab:zerofewshot} summarizes the macro-precision (P), macro-recall (R), and macro-F1 scores (F1) of different models on MCSD and MUStARD++ under zero-shot (ZS), few-shot (FS), 
and fine-tuning (FT) settings. 

\begin{table*}[t]
\centering
\caption{Macro-precision (P), macro-recall (R), and macro-F1 scores (F1)
on MCSD and MUStARD++ under zero-shot (ZS), few-shot (FS), and fine-tuning (FT) settings. For the FS results, we report
performance at $k=2$.}
\label{tab:zerofewshot}

\begin{tabular*}{\textwidth}{@{\extracolsep{\fill}}lccccccc}
\toprule
\textbf{Model} & \textbf{Setup} &
\multicolumn{3}{c}{\textbf{MCSD}} &
\multicolumn{3}{c}{\textbf{MUStARD++}} \\
\cmidrule(lr){3-5}
\cmidrule(lr){6-8}
& & \textbf{P} & \textbf{R} & \textbf{F1}
  & \textbf{P} & \textbf{R} & \textbf{F1} \\
\midrule

LLaMA 3
  & ZS  & 55.3 & 53.7 & 48.5 & 58.8 & 55.7 & 51.2 \\
8B
  & FS  & 64.3 & 56.0 & 46.7 & 64.1 & 58.0 & 52.6 \\
  & FT & 71.0 & 70.9 & 70.9 & 59.5 & 58.9 & 58.3 \\
\midrule

Qwen2-Audio
  & ZS  & 54.5 & 50.7 & 35.6 & 56.2 & 55.1 & 52.7 \\
7B
  & FS  & 53.9 & 53.9 & 53.7 & 54.2 & 54.0 & 53.6 \\
  & FT & 79.1 & 79.1 & 78.6 & 67.5 & 67.5 & 67.4 \\
\midrule

Qwen2.5-VL
  & ZS  & 27.0 & 50.0 & 35.0 & 58.9 & 50.6 & 35.9 \\
7B
  & FS  & 46.9 & 49.9 & 35.9 & 25.3 & 49.5 & 33.5 \\
  & FT & 67.9 & 68.0 & 67.9 & 61.5 & 61.2 & 61.0 \\
\midrule

Qwen2.5-Omni
  & ZS  & 57.4 & 57.2 & 56.3 & 62.5 & 62.3 & 62.2 \\
7B
  & FS  & 60.1 & 60.2 & 60.1 & 65.2 & 65.2 & 65.2 \\
  & FT & \textbf{80.8} & \textbf{80.8} & \textbf{80.3}
        & \textbf{70.5} & \textbf{70.3} & \textbf{70.1} \\
\bottomrule
\end{tabular*}
\end{table*}

Overall, fine-tuning with LoRA consistently leads to substantial improvements across all models and datasets. 
Qwen2-Audio and Qwen2.5-Omni reach FT F1 scores of 78.6\% and 80.3\% on MCSD, respectively, while on MUStARD++, Qwen2.5-Omni achieves the top FT F1 of 70.1\%, demonstrating that audio-text and trimodal integration can effectively capture sarcastic cues. 
On MCSD, LLaMA 3-8B shows a slight decrease from ZS to FS (ZS F1: 48.5\%, FS F1: 46.7\%), but FT dramatically boosts its F1 to 70.9\%. Similarly, on MUStARD++, LLaMA 3-8B FS performance increases slightly (ZS F1: 51.2\%, FS F1: 52.6\%), with FT improving F1 to 58.3\%. 
Qwen2-Audio benefits from few-shot prompting on MCSD (F1: 35.6\% to 53.7\%) but changes little on MUStARD++ (52.7\% to 53.6\%). Fine-tuning raises F1 to 78.6\% and 67.4\%, respectively.
Zero-shot performance is strongest for Qwen2.5-Omni (MCSD F1: 56.3\%, MUStARD++ F1: 62.2\%), illustrating that multimodal models can better capture sarcastic cues without additional examples. 
Among the models evaluated, Qwen2.5-Omni performs best across prompting settings.
Because the models differ in architecture as well as in accepted modalities, this
comparison does not isolate the effect of multimodality; Section~\ref{sec:modality}
addresses that directly.

Another noteworthy observation is the cross-dataset performance difference. LLaMA 3-8B performs relatively better on MUStARD++ than on MCSD
under zero-shot and few-shot settings, whereas the Qwen models show stronger performance across both datasets. This pattern is consistent with a possible benefit of broader multilingual pretraining, including greater exposure to Chinese data. However, because the two datasets also differ in domain, speaking style, and cultural context, these results should not be interpreted as isolating the effect of pretraining language coverage alone. 

\subsection{Modality Attribution Analysis}
\label{sec:modality}

\begin{figure*}[htbp]
    \centering
    \includegraphics[width=\textwidth]{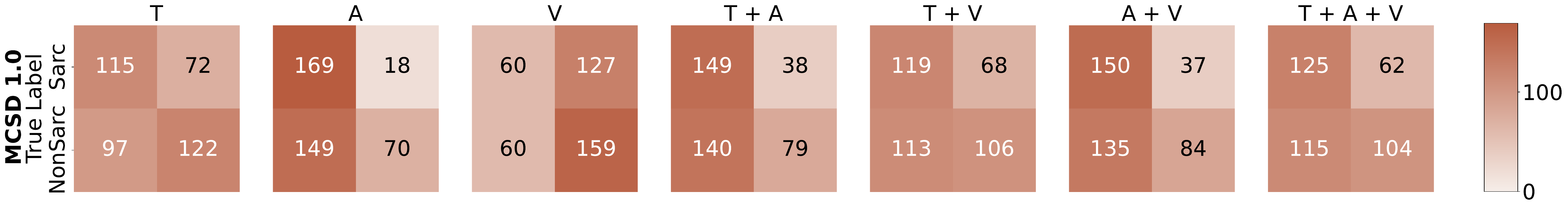}\\[0.5em]
    \includegraphics[width=\textwidth]{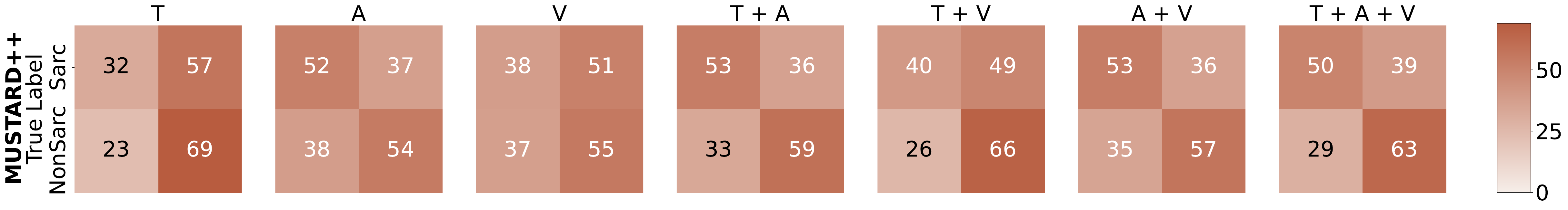}
    \caption{
        Confusion matrices for different modality configurations on
        (top) MCSD and (bottom) MUStARD++.
    }
    \label{fig:confusion_matrices}
\end{figure*}

To understand how different modalities contribute to sarcasm detection, we conduct controlled experiments with Qwen2.5-Omni by restricting the model's access to specific modality inputs.
Table~\ref{tab:qwen25omni_zs_modality} reports the zero-shot performance of Qwen2.5-Omni under different modality configurations.
Figure~\ref{fig:confusion_matrices} shows the corresponding confusion matrices, providing a fine-grained analysis of modality-specific error patterns.

\begin{table*}[ht]
\centering
\small
\caption{Zero-shot performance of Qwen2.5-Omni under different modality configurations on MCSD and MUStARD++. T, A, and V denote text, audio, and visual inputs, respectively. The text-only condition uses reference transcripts. The unimodal, bimodal, and trimodal settings provide a controlled analysis of modality contribution and reduced-modality availability.}
\label{tab:qwen25omni_zs_modality}
\begin{tabular*}{\textwidth}{@{\extracolsep{\fill}}lcccccc}
\toprule
\textbf{Setup} &
\multicolumn{3}{c}{\textbf{MCSD}} &
\multicolumn{3}{c}{\textbf{MUStARD++}} \\
\cmidrule(lr){2-4} \cmidrule(lr){5-7}
 & \textbf{P} & \textbf{R} & \textbf{F1} & \textbf{P} & \textbf{R} & \textbf{F1} \\
\midrule
T
  & 58.6 & 58.6 & \textbf{58.4}
  & 56.5 & 55.5 & 53.9 \\
A
  & \textbf{66.4} & \textbf{61.2} & 56.3
  & 58.6 & 58.6 & 58.6 \\
V
  & 52.8 & 52.3 & 51.0
  & 51.3 & 51.2 & 51.0 \\
\midrule
T+A
  & 59.5 & 57.9 & 54.8
  & 61.9 & 61.8 & 61.8 \\
T+V
  & 56.1 & 56.0 & 55.4
  & 59.0 & 58.3 & 57.7 \\
A+V
  & 61.0 & 59.3 & 56.5
  & 60.8 & 60.8 & 60.8 \\
\midrule
T+A+V
  & 57.4 & 57.2 & 56.3
  & \textbf{62.5} & \textbf{62.3} & \textbf{62.2} \\
\bottomrule
\end{tabular*}
\end{table*}

The relative effectiveness of individual modalities is dataset-dependent. On MUStARD++, audio (A) achieves the strongest unimodal performance, whereas on MCSD, text (T) performs best among the three unimodal configurations. The text-only condition uses reference transcripts, whereas the audio condition receives the speech signal directly; because Qwen2.5-Omni can transcribe speech, the audio-only setting may also provide access to lexical information. Under our frame-based visual representation, vision yields comparatively lower performance, particularly on MUStARD++. Our visual pipeline samples frames and does not preserve temporal facial expression, gaze, or gesture, so weak visual performance here reflects the representation rather than the informativeness of visual cues. Gao et al. \cite{gao2025multimodal} report a similar effect, with a higher frame rate yielding 45.27\% F1.
%
The effect of combining modalities is configuration- and dataset-dependent. On MCSD, A+V achieves the strongest bimodal performance (56.5\% F1), although it remains below the text-only condition (58.4\%). On MUStARD++, T+A is the strongest bimodal configuration (61.8\%), followed by A+V (60.8\%), with both improving over their corresponding unimodal components. Thus, no single bimodal combination is consistently optimal across datasets.
The full-modality configuration also shows contrasting effects across the two datasets. On MUStARD++, T+A+V achieves the best overall performance at 62.2\% F1, suggesting that the three modalities provide complementary information in this setting. On MCSD, however, T+A+V reaches 56.3\% F1 and does not outperform the text-only condition (58.4\%). These results show that adding modalities does not lead to monotonic performance gains and that their effectiveness depends on both the dataset and the modalities already available.

\begin{figure*}[htbp]
    \centering
    \includegraphics[width=0.8\textwidth]{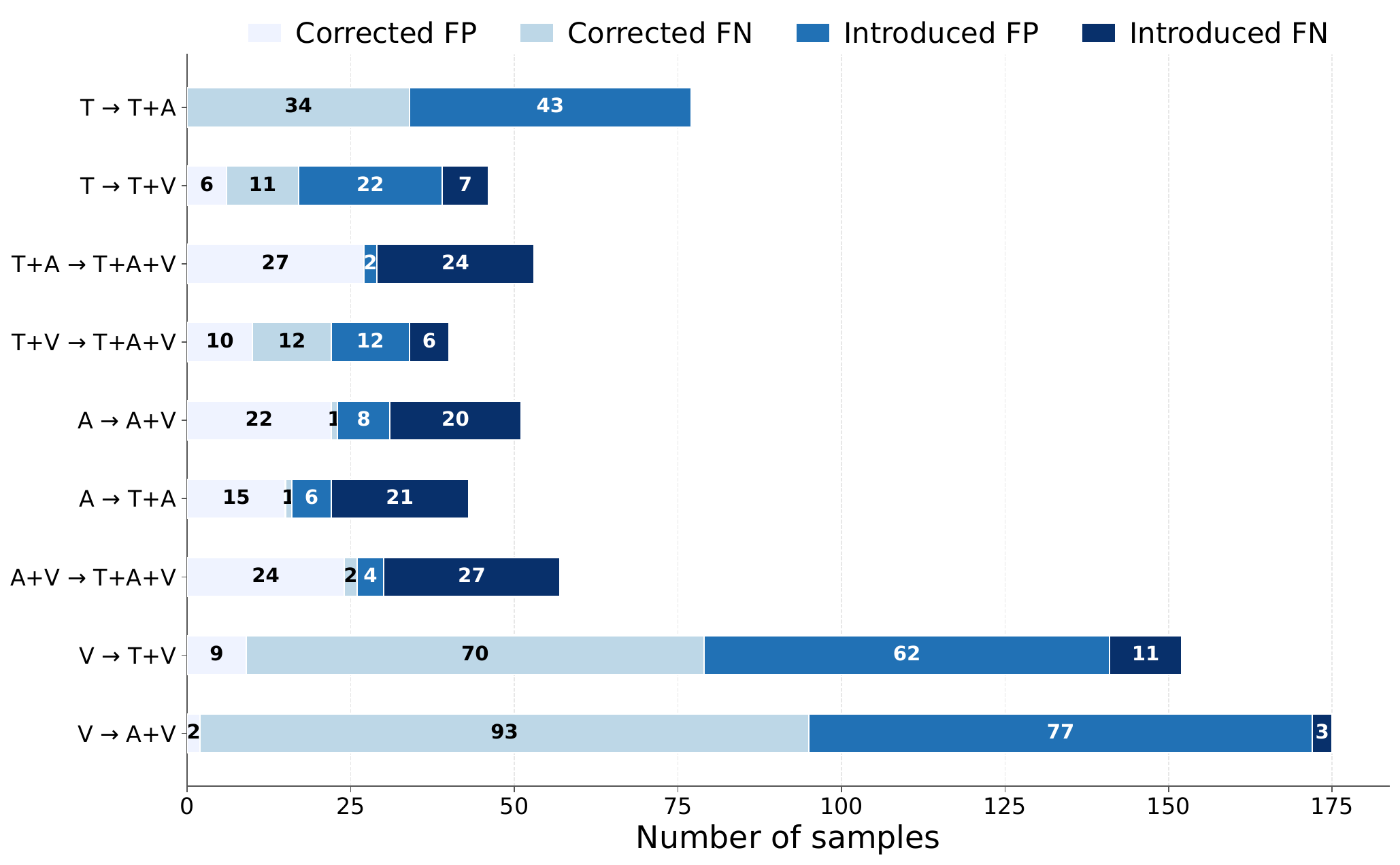}\\[0.5em]
    \includegraphics[width=0.95\textwidth]{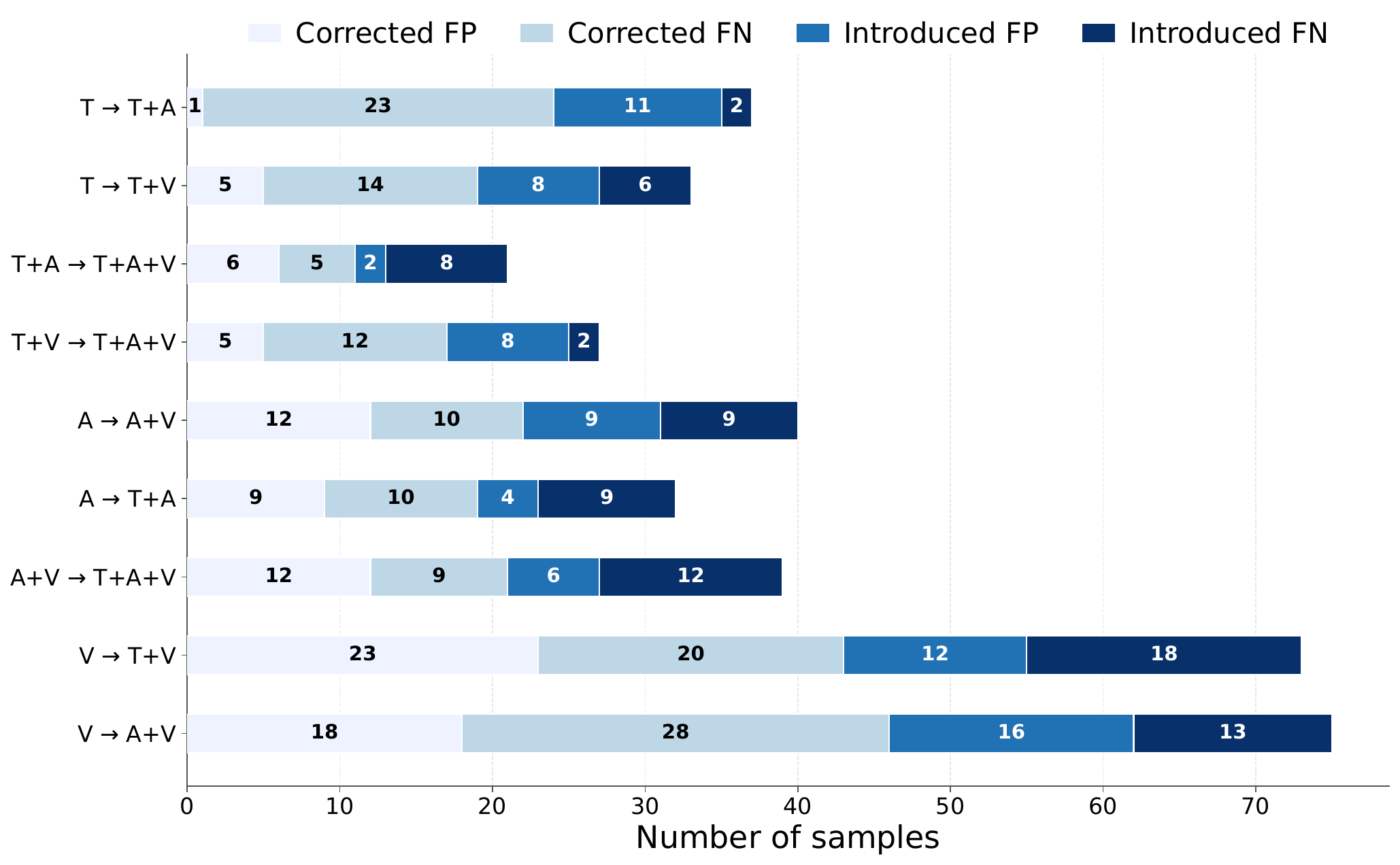}
    \caption{Errors corrected and newly introduced by adding modalities as input to Qwen2.5-Omni on MCSD (top) and MUStARD++ (bottom). T, A, and V denote text, audio, and visual inputs, respectively; ``$\rightarrow$'' denotes a transition from the modality configuration on the left to that on the right.}
    \label{fig:error}
\end{figure*}

To move beyond aggregate performance scores, Figure~\ref{fig:error} examines how adding each modality changes individual predictions. For each transition, such as T$\rightarrow$TA, we report the number of false positives (FPs) and false negatives (FNs) that are corrected, as well as the new errors introduced by the additional modality. Here, T$\rightarrow$TA compares text-only and text-audio predictions on the same test instances.

\paragraph{What errors does each modality correct?}
Across both MCSD and MUStARD++, adding audio to text primarily corrects FNs rather than FPs. On MCSD, \textsc{T$\rightarrow$TA} corrects 34 FNs and no FPs, while on MUStARD++ it corrects 23 FNs and only 1 FP. This suggests that acoustic information mainly enhances recall, helping the model identify sarcasm that is weakly signaled or ambiguous in the reference transcript but expressed through the speech signal.

Visual information produces a similar, although weaker, pattern. On MCSD, \textsc{T$\rightarrow$TV} corrects 11 FNs versus 6 FPs, while on MUStARD++ it corrects 14 FNs versus 5 FPs. Thus, when text is the base modality, both audio and vision primarily help recover missed sarcasm, with audio showing the clearest FN-correction effect.

A different behavior emerges when augmenting audio-only input. On MCSD, adding either text or vision overwhelmingly corrects FPs: \textsc{A$\rightarrow$TA} corrects 15 FPs versus 1 FN, while \textsc{A$\rightarrow$AV} corrects 22 FPs versus 1 FN. On MUStARD++, however, the effects are more balanced: \textsc{A$\rightarrow$TA} corrects 9 FPs and 10 FNs, whereas \textsc{A$\rightarrow$AV} corrects 12 FPs and 10 FNs. 
Chen et al. \cite{chen2026hear} show that MLLMs respond to a stereotype of expressive prosody, raised pitch and irregular pausing, which diverges from the acoustic signatures of sarcasm in both languages and produces false positives. The MCSD results fit that account.
Textual and visual information can therefore provide grounding signals that suppress audio-induced sarcasm hypotheses, particularly on MCSD.

Overall, audio and vision mainly help recover sarcasm missed by the text-only model. In contrast, text and vision primarily reduce false positives produced by the audio-only model on MCSD, while their effects on MUStARD++ are more evenly distributed across FPs and FNs.

When moving from bimodal to trimodal settings, the effect depends on the existing modality pair. On MCSD, \textsc{TA$\rightarrow$TAV} corrects 27 FPs versus 2 FNs, while \textsc{AV$\rightarrow$TAV} corrects 24 FPs versus 2 FNs, indicating a strong reduction in spurious sarcasm predictions. However, \textsc{TV$\rightarrow$TAV} corrects a more balanced set of errors (10 FPs and 12 FNs). 
On MUStARD++, the corrections are more evenly split. Adding vision to T+A corrects 6 FPs and 5 FNs, and adding text to A+V corrects 12 FPs and 9 FNs. Adding audio to T+V is the exception, correcting 5 FPs and 12 FNs.
Full multimodal fusion therefore does not serve a single role; its effect depends on both the dataset and the modalities already available.

\paragraph{What new errors does each modality introduce?}
While additional modalities correct many errors, they also disrupt previously correct predictions. The types of newly introduced errors again depend on the base modality and the added signal.

When augmenting text with audio or visual information, most newly introduced errors are FPs. On MCSD, \textsc{T$\rightarrow$TA} introduces 43 FPs and no FNs, while \textsc{T$\rightarrow$TV} introduces 22 FPs versus 7 FNs. The same tendency appears on MUStARD++, where \textsc{T$\rightarrow$TA} introduces 11 FPs versus 2 FNs and \textsc{T$\rightarrow$TV} introduces 8 FPs versus 6 FNs. These results indicate that non-textual information can override correct textual interpretations and trigger spurious sarcasm predictions. Expressive acoustic patterns or salient visual behavior may be interpreted as sarcastic even when the lexical content is sincere.

Conversely, adding text or vision to audio-based models tends to introduce FNs. On MCSD, \textsc{A$\rightarrow$TA} introduces 21 FNs versus 6 FPs, while \textsc{A$\rightarrow$AV} introduces 20 FNs versus 8 FPs. On MUStARD++, \textsc{A$\rightarrow$TA} introduces 9 FNs versus 4 FPs, whereas \textsc{A$\rightarrow$AV} introduces an equal number of FPs and FNs (9 each). Grounding signals can therefore make the model more conservative, sometimes suppressing genuine sarcasm that is conveyed strongly through acoustic information.

In bimodal-to-trimodal transitions, the direction of the trade-off depends on the starting pair. Adding the third modality to T+A or A+V mainly introduces FNs: on MCSD, \textsc{TA$\rightarrow$TAV} and \textsc{AV$\rightarrow$TAV} introduce 24 and 27 FNs, respectively, and the corresponding values on MUStARD++ are 8 and 12. By contrast, \textsc{TV$\rightarrow$TAV} mainly introduces FPs on both datasets (12 FP vs.\ 6 FN on MCSD; 8 FP vs.\ 2 FN on MUStARD++). Thus, cross-modal integration may either suppress valid sarcasm or encourage spurious predictions, depending on which modality is added.

Overall, these results show a trade-off in multimodal sarcasm detection. Adding a modality can correct errors under the existing configuration while also introducing new ones, with the balance depending on the base modality, the added signal, and the dataset. McNemar tests \cite{mcnemar1947note} show that none of the modality transitions produces a significant overall change after Holm correction, but class-stratified tests show substantial shifts in error type, particularly on MCSD. For example, adding audio to text on MCSD significantly reduces false negatives while significantly increasing false positives (both Holm-adjusted \(p<.001\)). 


\subsection{Effect of $k$-sample on Few-shot Performance}

\begin{figure}[h!]
    \centering
    \includegraphics[width=0.8\linewidth]{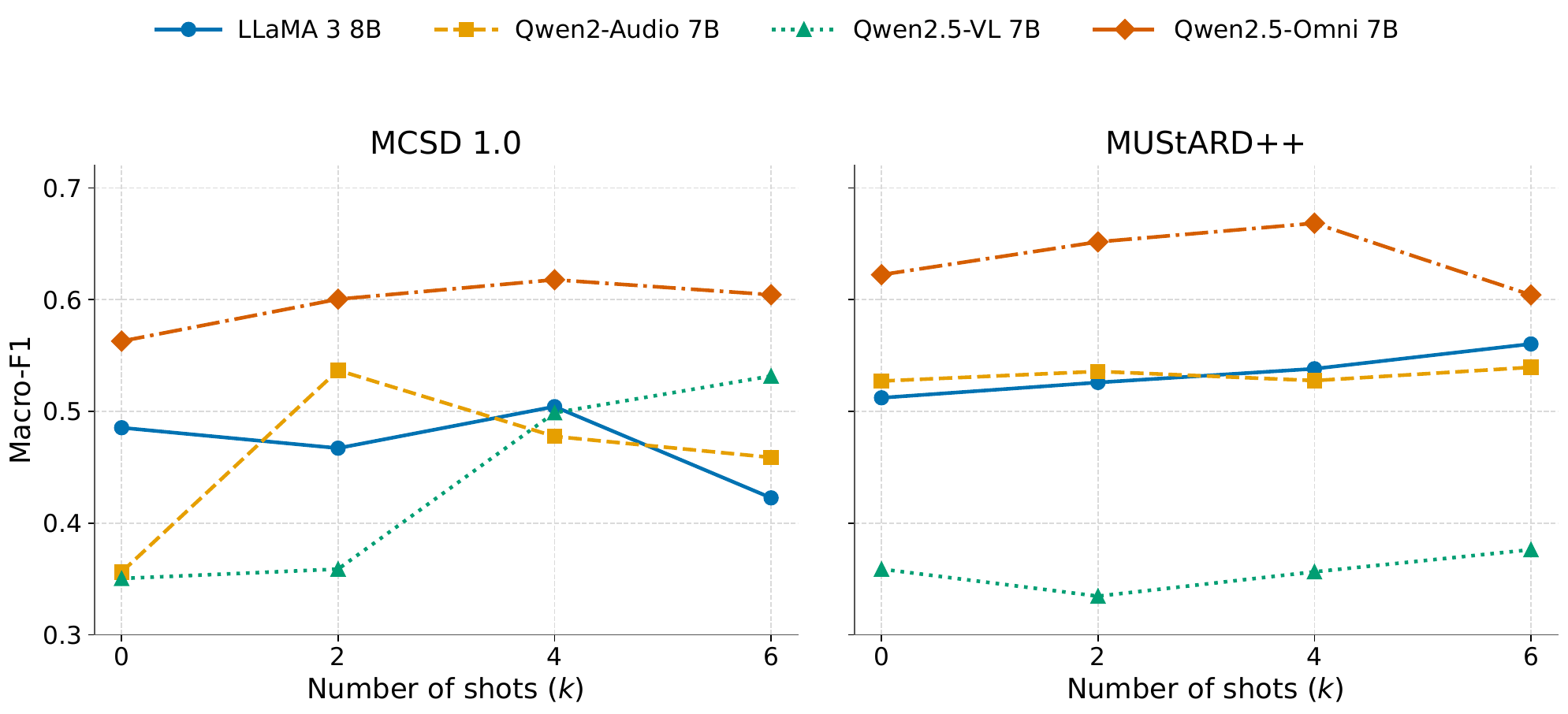}
    \caption{Few-shot performance comparison of MLLMs under different $k$ values.}
    \label{fig:fewshot}
\end{figure}

To investigate in-context learning abilities of MLLMs, we analyze the effect of varying the number of few-shot examples (Figure~\ref{fig:fewshot}).

LLaMA 3-8B improves on MUStARD++ from an F1 of 51.2\% at 0-shot to 56.0\% at 6-shot. On MCSD, performance fluctuates and peaks at 50.4\% with 4-shot, indicating that few-shot prompting provides inconsistent benefits for the text-only model.
Qwen2-Audio-7B shows little improvement on MUStARD++, increasing from 52.7\% to 54.0\%. On MCSD, F1 rises from 35.6\% to 53.7\% at 2-shot before declining, suggesting that a small number of examples can help, although the gains are not stable.
Qwen2.5-VL-7B improves only slightly on MUStARD++, from 35.9\% to 37.6\%. On MCSD, however, F1 increases from 35.0\% to 53.2\% at 6-shot, showing that visual-language models can benefit considerably from additional demonstrations.
Qwen2.5-Omni-7B achieves the strongest few-shot results, reaching 66.9\% on MUStARD++ and 61.8\% on MCSD at 4-shot. Performance declines slightly at 6-shot, indicating that few-shot gains are non-monotonic and that using more examples does not necessarily yield better performance.


\subsection{Effect of Training Data Size on LoRA Fine-Tuning}

\begin{figure}[h!]
    \centering
    \includegraphics[width=0.8\linewidth]{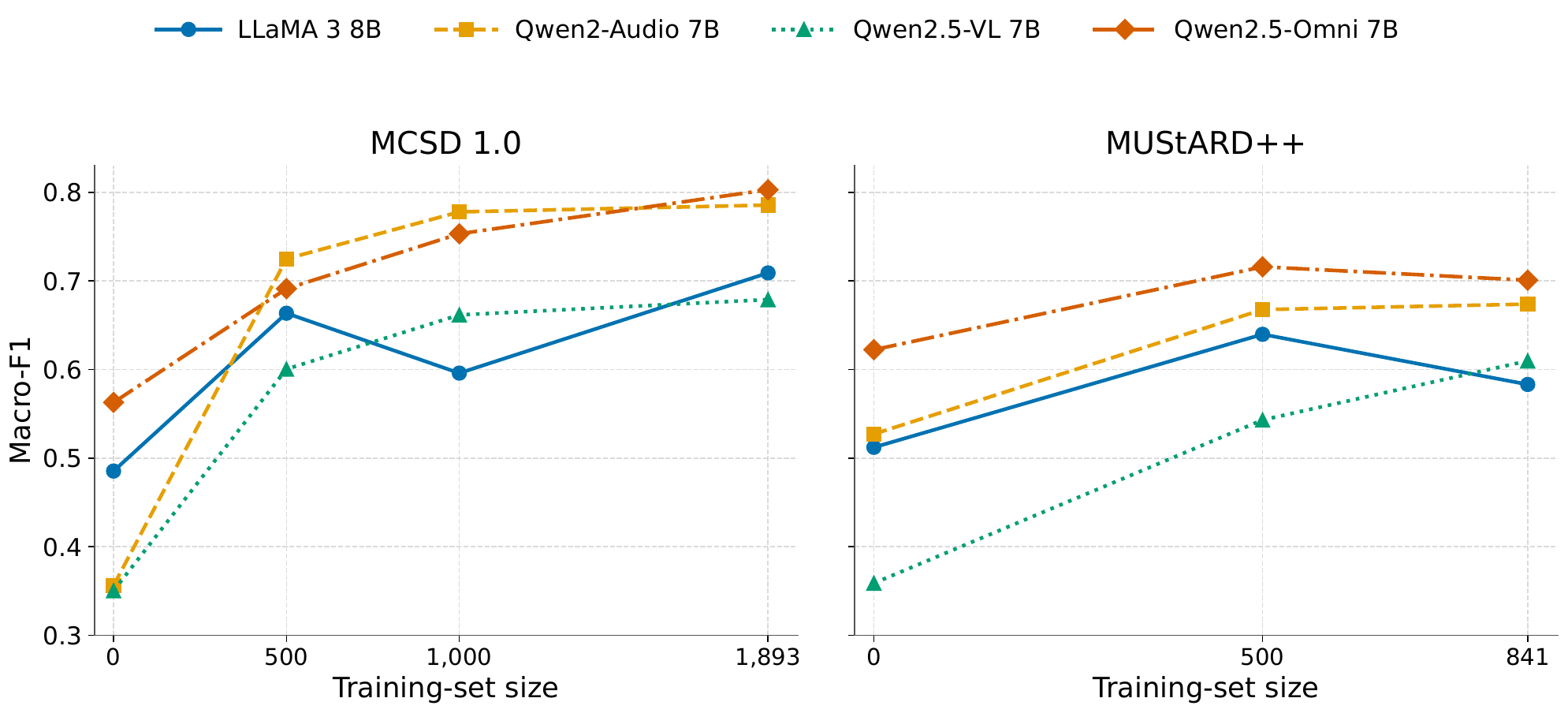}
    \caption{Fine-tuning performance comparison of large-scale models under different train sizes.}
    \label{fig:train_size}
\end{figure}

We study the impact of different training set sizes on the fine-tuning performance of various MLLMs using LoRA. Figure~\ref{fig:train_size} summarizes model performance across different training sizes, with MCSD evaluated at 0, 500, 1000, and all 1,893 available samples, and MUStARD++ evaluated at 0, 500, and all 841 samples.

Overall, increasing the training size leads to a consistent improvement in F1 performance for most models. On MCSD, Qwen2.5-Omni-7B achieves the highest F1 score, reaching 80.3\% with all 1,893 training samples, followed closely by Qwen2-Audio-7B. LLaMA 3-8B and Qwen2.5-VL-7B show moderate gains. 
%
On MUStARD++, the trend is less uniform due to the presence of small data effects.
While Qwen2-Audio-7B improves with more data, LLaMA 3-8B and Qwen2.5-Omni-7B achieve the best performance at 500 samples but slightly decrease at 841 samples, which may reflect small-sample variability or dataset-specific effects.
Qwen2.5-VL-7B shows steady improvement, although overall performance is lower compared to other models. 
These results indicate that the performance gains from LoRA fine-tuning are model- and dataset-dependent, and careful selection of training size is crucial for maximizing the benefit of parameter-efficient fine-tuning on MLLMs.

\subsection{Unimodal and Multimodal Model Performance}
\begin{figure}[h!]
    \centering
    \includegraphics[width=0.8\linewidth]{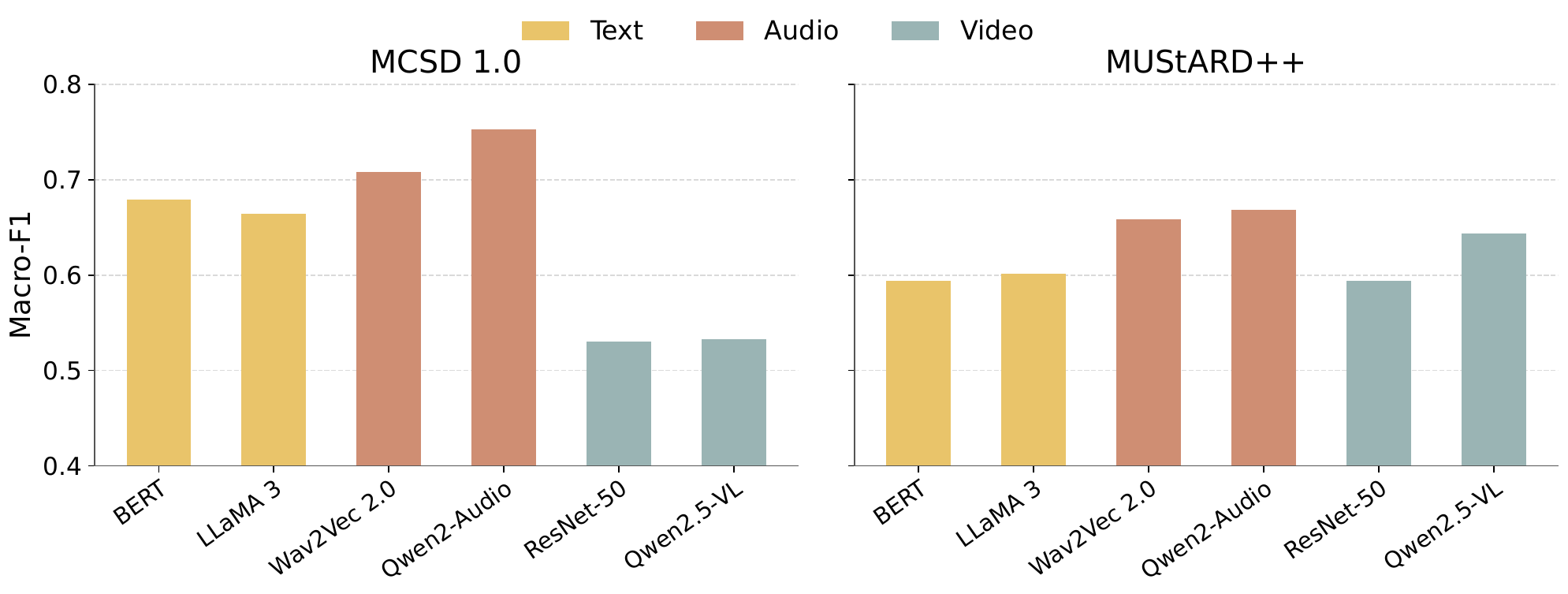}
    \caption{Macro-F1 score comparison for different models on MCSD and MUStARD++ across unimodal settings.}
    \label{fig:f1_single_modality}
\end{figure}

We compare unimodal and multimodal performances on MCSD and MUStARD++, with results reported in Figure~\ref{fig:f1_single_modality} and Table~\ref{tab:results_plm}.


\textbf{Unimodal performance:} 
In unimodal settings, audio-based models consistently outperform text- or vision-based ones. On MCSD, Qwen2-Audio reaches the highest F1 score (75.3\%). On MUStARD++, Qwen2-Audio achieves the strongest performance with 66.8\% F1. Visual models show lower performance under the frame-based representation adopted in our experiments.
As Qwen2-Audio is trained on speech and text, its audio representations may encode lexical and acoustic content, so this comparison does not isolate prosody.

\textbf{Bimodal fusion:}
The effectiveness of bimodal fusion varies across datasets and modality combinations. On MUStARD++, Large (A+V) achieves the best bimodal F1-score of 71.2\%, outperforming both audio-only (66.8\%) and vision-only (64.4\%) models. Large (T+A) also reaches 70.3\%, whereas T+V provides only limited gains (64.6\%). On MCSD, Base (T+A) produces the strongest bimodal result at 74.7\% F1, followed by Large (T+A) at 73.5\%. However, neither surpasses the best unimodal audio model (75.3\%), suggesting that modality fusion does not consistently improve upon strong unimodal representations.

\begin{table*}[ht]
\centering
\caption{Precision (P), Recall (R), and F1-scores (F1) in bimodal and trimodal settings, using models as feature extractors. Base (T): BERT; Large (T): LLaMA 3; Base (A): Wav2Vec 2.0; Large (A): Qwen2-Audio;  Base (V): ResNet-50; Large (V): Qwen2.5-VL.}
\label{tab:results_plm}
\begin{tabular*}{\textwidth}{@{\extracolsep{\fill}}lcccccc}
\toprule
\textbf{Model} &
\multicolumn{3}{c}{\textbf{MCSD}} &
\multicolumn{3}{c}{\textbf{MUStARD++}} \\
\cmidrule(lr){2-4} \cmidrule(lr){5-7}
& \textbf{P} & \textbf{R} & \textbf{F1} & \textbf{P} & \textbf{R} & \textbf{F1} \\
\midrule
Base (T+A)
    & 74.8 & \textbf{74.9} & 74.7
    & 67.1 & 66.8 & 66.5 \\
Large (T+A)
    & 73.6 & 73.8 & 73.5
    & 70.4 & 70.3 & 70.3 \\
\midrule
Base (T+V)
    & 66.7 & 66.6 & 66.6
    & 64.7 & 64.6 & 64.6 \\
Large (T+V)
    & 67.4 & 67.5 & 67.4
    & 64.7 & 64.7 & 64.6 \\
\midrule
Base (A+V)
    & 69.1 & 69.0 & 69.0
    & 66.3 & 66.3 & 66.3 \\
Large (A+V)
    & 71.6 & 71.7 & 71.5
    & 71.6 & 71.4 & 71.2 \\
\midrule
Base (T+A+V)
    & 72.0 & 71.8 & 71.7
    & 66.4 & 66.4 & 66.3 \\
Large (T+A+V)
    & \textbf{74.8} & 74.8 & \textbf{74.7}
    & \textbf{73.5} & \textbf{73.4} & \textbf{73.3} \\
\bottomrule
\end{tabular*}
\end{table*}

\textbf{Trimodal fusion:} 
Adding all three modalities (T+A+V) produces contrasting results across the two datasets. On MCSD, Large (T+A+V) reaches 74.7\% F1, matching the strongest bimodal result but remaining slightly below the audio-only model (75.3\%). On MUStARD++, however, Large (T+A+V) achieves the best overall performance at 73.3\%, exceeding Large (A+V) by 2.1 percentage points. 
Compared with existing task-specific systems, our gated fusion over MLLM features does not achieve the highest reported performance on either dataset. However, these results are not directly comparable because prior work uses different evaluation settings. Overall, direct LoRA fine-tuning of Qwen2.5-Omni reaches 80.3\% F1 on MCSD, higher than the 76.64\% reported by the original MCSD baseline, although the evaluation protocols differ \cite{gao2025multimodal}.

\section{Conclusion}

In this work, we systematically evaluated LLMs and MLLMs for spoken sarcasm detection on English and Chinese datasets. We considered two complementary settings: directly prompting or fine-tuning large models for classification, and using pretrained models as feature extractors within a collaborative gating framework. Qwen2.5-Omni achieves the strongest zero-shot performance, while LoRA fine-tuning substantially improves performance across models, highlighting the importance of task-specific adaptation.

The role of audio differs across modeling settings. In the feature-extraction experiments, audio features provide the strongest unimodal performance on both datasets. In the zero-shot Qwen2.5-Omni setting, however, text-only input based on reference transcripts outperforms audio-only input on MCSD. Error-transition analysis further shows that adding acoustic information can recover sarcasm missed from text alone but can also introduce false positives, indicating that its contribution depends on how multimodal information is represented and integrated.

The benefit of multimodal integration is also dataset-dependent. In the feature-extraction setting, trimodal fusion achieves the best overall performance on MUStARD++, whereas on MCSD it remains slightly below the strongest audio-only model. These findings suggest that effective sarcasm detection requires selective, dataset-sensitive multimodal integration rather than simply combining all available modalities. Future work should explore dynamic fusion strategies, broader multilingual evaluation, and more interpretable models of sarcastic intent.

\section{Limitations and Future Work}
Due to the computational cost of training large multimodal models, the fine-tuning and feature-fusion experiments were conducted as single runs. Therefore, the small performance differences should be interpreted with caution, particularly on the relatively small MUStARD++ test set. Future work could further investigate post-training strategies for adapting MLLMs to multimodal sarcasm detection, some of which have already begun to be explored in \cite{li2026sarcasmminer}.
\section{Ethics and Data Statement}
MUStARD++ and MCSD consist of clips from publicly broadcast sitcoms and stand-up comedy featuring identifiable performers. We use both datasets in accordance with their terms of release. We do not redistribute the audiovisual material and perform no identity-related analysis.

\bibliographystyle{splncs04}
\bibliography{custom}

\end{document}